\documentclass{article}
\usepackage{ijcai20}

\usepackage{times}
\usepackage{soul}
\usepackage{url}
\usepackage[hidelinks]{hyperref}
\usepackage[utf8]{inputenc}
\usepackage[small]{caption}
\usepackage{graphicx}
\usepackage{amsmath}
\usepackage{amsthm}
\usepackage{booktabs}
\usepackage{algorithm}
\usepackage{algorithmic}
\usepackage{cite}
\usepackage{amsmath}
\usepackage{amsfonts}
\usepackage{graphicx}
\usepackage{multirow}
\usepackage[round]{natbib}
\newtheorem{theorem}{Theorem}
\newtheorem{proposition}{Proposition}

\newtheorem{definition}{Definition}

\usepackage{algorithm}
\usepackage{color}
\usepackage{xcolor}
\usepackage{algorithmic}
\numberwithin{figure}{section}

\newcommand{\bx}{\mathbf{x}}

\newcommand{\bI}{\mathbf{I}}

\newcommand{\bP}{\mathbf{P}}

\newcommand{\bU}{\mathbf{U}}

\newcommand{\bX}{\mathbf{X}}

\newcommand{\bbR}{\mathbb{R}}

\usepackage{mathrsfs}

\title{SymNMF-Net for The Symmetric NMF Problem}

\author{
Mingjie Li$^1$
\and
Hao Kong$^1$\and
Zhouchen Lin$^{1}$
\affiliations
$^1$Peking University
}

\begin{document}

\maketitle

\begin{abstract}
	Recently, many works have demonstrated that Symmetric Non-negative Matrix Factorization~(SymNMF) enjoys a great superiority for various clustering tasks.
	Although the state-of-the-art algorithms for SymNMF perform well on synthetic data, they cannot consistently obtain satisfactory results with desirable properties and may fail on real-world tasks like clustering. Considering the flexibility and strong representation ability of the neural network, in this paper, we propose a neural network called SymNMF-Net for the Symmetric NMF problem to overcome the shortcomings of traditional optimization algorithms.
	Each block of SymNMF-Net is a differentiable architecture with an inversion layer, a linear layer and ReLU, which are inspired by a traditional update scheme for SymNMF. We show that the inference of each block corresponds to a single iteration of the optimization. Furthermore, we analyze the constraints of the inversion layer to ensure the output stability of the network to a certain extent. Empirical results on real-world datasets demonstrate the superiority of our SymNMF-Net and confirm the sufficiency of our theoretical analysis.

\end{abstract}

\section{Introductions}
Non-negative Matrix Factorization~(NMF) plays an important role in feature extraction task due to its powerful linear representation ability~\citep{face2,shahnaz2006document,probcluster,kang2014nmf,kuang2015symnmf,zhu2018dropping}. The Classical NMF problem can be formulated as the following:

Given a data matrix $\mathbf{X}\in\mathbb{R}^{m\times n}$, NMF is referred to finding non-negative matrices $\mathbf{A}\in\mathbb{R}^{m\times r}$ and $\mathbf{B}\in\mathbb{R}^{n\times r}$ with $r\ll \min\{m,n\}$ to optimize the following:
\begin{equation}
\label{prob:nmf}
\min_{\mathbf{A}\geq 0, \mathbf{B}\geq 0} \frac{1}{2}\|\mathbf{X}-\mathbf{A}\mathbf{B}^\top\|^2_F.
\end{equation}
In some specific tasks such as image or text clustering, $\mathbf{X}\in\mathbb{R}^{n\times n}$ can be chosen as a symmetric relational matrix~\citep{zhu2018dropping,probcluster}. Accordingly, Eq.~(\ref{prob:nmf}) can be reformulated as the following:
\begin{equation}
\label{prob:snmf}
\begin{aligned}
\min_{\mathbf{U}\in\mathbb{R}^{n\times r}}\  \frac{1}{2}\|\mathbf{X-UU}^\top\|_F^2,\quad s.t.\ \mathbf{U}\geq 0,
\end{aligned}
\end{equation}
where $\mathbf{U}\mathbf{U}^\top$ corresponds to a symmetric matrix with rank $r$. 
Many related works~\citep{kuang2015symnmf,probcluster,ding2005equivalence} have analyzed the superiority of SymNMF over other methods in solving clustering problems.
In particular,~\cite{ding2005equivalence} systematically prove that SymNMF is equivalent to the kernel $k$-means and Laplacian-based spectral clustering. 
In addition to solving Eq.~(\ref{prob:snmf}) Projected Gradient Descent~(PGD) directly, \cite{kuang2015symnmf} introduce an auxiliary variable $\mathbf{V}$ as another factor matrix, and add $\frac{\lambda}{2}\|\mathbf{W}-\mathbf{V}\|_F^2$ to the objective function as a penalty term, formulated as follow:
\begin{equation}
\label{prob:drop-snmf}
\begin{aligned}
&\min_{\mathbf{U}\in\mathbb{R}^{n\times r}, \mathbf{V}\in\mathbb{R}^{n \times r}}\  \frac{1}{2}\|\mathbf{X-WV}^\top\|_F^2+\frac{\lambda}{2}\|\mathbf{W-V}\|_F^2,\\
&\quad\qquad s.t.\qquad\ \mathbf{W}\geq 0, \ \mathbf{V}\geq 0.
\end{aligned}
\end{equation}
\par
The most successful algorithms based on the above formulation are the Symmetric Alternative Non-negative Least Square (SymANLS) and the Symmetric Hierarchical Least Square (SymHALS) which are proposed by \cite{zhu2018dropping}.

Although the classical approaches can reach to some local minima quickly, they can randomly approximate the non-negative factorization with respect to the initialization. Some special attributes are not considered in the algorithm which makes it hard to produce	the results with desired properties. On this account, the classical algorithms often fail on some real-world tasks such as clustering.
\par	
Due to the above defects for the classical algorithm, we deem that the \textbf{design} of a practical factorization algorithm should be data-driven or target-driven. Thus, the factor matrices can extract more effective intrinsic information from data, which may lead to better and more consistent performance.
%

Deep Neural Networks~(DNNs) have now achieved great success in many machine learning tasks.
On the one hand, back-propagation makes the training process driven by an objective function which is usually designed by specific requirements. 
On the other hand, training DNN with gradient descent or its variants can ensure the network reach its local minima with proper learning rate even the structure is non-convex in most cases especially with perturbed gradient descent method~\citep{jin2017escape}. Inspired by these observations, we want to use a DNN to construct a data-driven architecture for the SymNMF problem.
In general, it is quite difficult for DNN to solve many traditional mathematical models directly, such as SymNMF, because the constraints of these problems are hard to be structured as commonly used neural layers or formulated as a part of loss functions. 
%

In this paper, inspired by a straightforward update scheme for problem (\ref{prob:drop-snmf}), we propose a new network structure to solve the SymNMF problem. Each block contains three layers sequentially: the inversion layer, linear layer, and ReLU layer. 
Moreover, in order to ensure the output stability of the inversion layer, we analyze the sufficient lower bound for the penalty parameter $\lambda$, which is also a parameter in our network.
In summary, our main contributions include:
\begin{itemize}
	\item 
	We parameterize the optimization iteration of a numerical updating scheme and design SymNMF-Net for solving the symNMF problem~(\ref{prob:drop-snmf}). Each block of SymNMF-Net is a differentiable architecture with an inversion layer, a linear layer and ReLU.
	\item 
	We analyze the lower bound for the learnable parameter $\lambda$ to ensure the output stability of the inversion layer. We also show that our SymNMF-Net shares some critical points with traditional methods under certain conditions.
	\item
	We conduct experiments on numerical comparison and graph clustering to validate the effectiveness of our SymNMF-Net and our theoretical bound for $\lambda$.
\end{itemize}
\subsection{Related Work}

Differentiable programming, or learning-based optimization, has achieved great success in many tasks. For example, \cite{kmeansnet} combine K-means with neural network for clustering. Many works~\citep{sprechmann2015learning,liu2018alista,chen2018theoretical} solve the LASSO problem via a recurrent neural network~(RNN). \cite{zhou2018sc2net} involve long short time memory~(LSTM) for sparse coding. ADMM-Net~\citep{sun2016deep} and Differentiable Linear ADMM~\citep{dladmm} unfold the ADMM algorithm to solve the problems with linear constraints. The empirical experiments of these works demonstrate their improvements over the original optimization algorithms. However, their frameworks cannot be used directly to solve NMF problems.

\section{The SymNMF-Net}
\label{struct}
A straight-forward way to overcome the SymNMF problem~(\ref{prob:drop-snmf}) is implementing the following schemes iteratively,
\begin{equation}
\label{tradi_update}
\left\{
\begin{array}{l}
\mathbf{W}_k = \max\{(\mathbf{X}+\lambda\mathbf{I}_n)\mathbf{V}_{k-1}(\mathbf{V}_{k-1}^\top\mathbf{V}_{k-1}+\lambda \mathbf{I}_r)^{-1},0\},\\
\mathbf{V}_k = \max\{(\mathbf{X}+\lambda\mathbf{I}_n)\mathbf{W}_{k}(\mathbf{W}_k^\top\mathbf{V}_k+\lambda \mathbf{I}_r)^{-1},0\}.
\end{array}
\right.
\end{equation}
\par
Unfortunately, the above method may be trapped somehow before the local minima. Owing to that fact, classical algorithms often involve complex methods to tackle the problem. However, if we unfold this diagram to a neural network and use gradient descent or noisy gradient method via reconstruction loss to train it, we may easily pull it from the saddle points in most cases and obtain local minima of the non-convex problem with the guarantee of the gradient method.

\begin{figure}[t]
	\centering
	\includegraphics[width=0.5\textwidth]{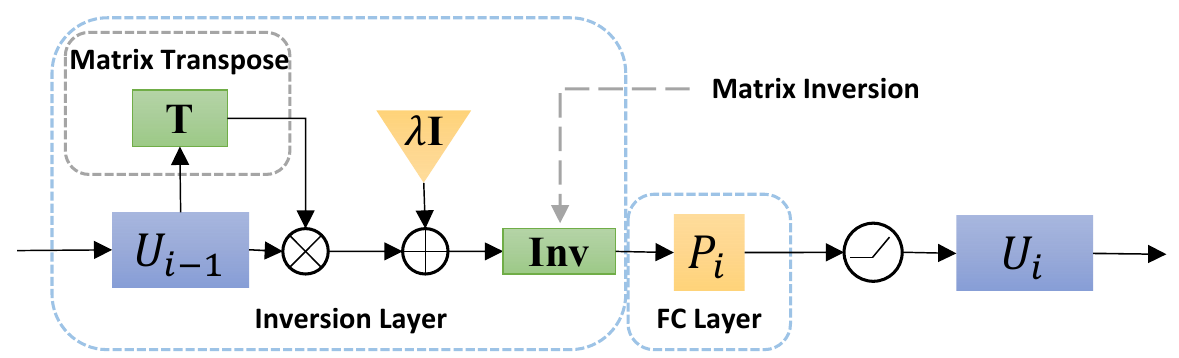}
	\caption{The structure of the $i$-th SymNMF Block with input $\bU_{i-1}$ and output $\bU_i$. $\bP_i$ and $\lambda$ are learnable parameters which we use back-propagation to update them.}
	\label{fig:SymNMF-Net-block}
\end{figure}

Motivated by such insights, we unfold the above algorithm into a neural network by adding an inversion layer into the network. Inputting a factor matrix $\mathbf{U}$, the output inversion layer can be formulated as $(\mathbf{U}^\top\mathbf{U}+\lambda \mathbf{I}_r)^{-1}$.
We multiply the input factor matrix with its transpose and then add a weighted identity matrix $\lambda\mathbf{I}$ to circumvent the ill-posedness of $\mathbf{U}^\top\mathbf{U}$. Finally, we use a matrix inversion to get the output of the inversion layer~(the left part in Fig.~\ref{fig:SymNMF-Net-block}). Moreover, considering the condition number of the inversion layer\footnote{$\sigma_1(\cdot)$ and $\sigma_n(\cdot)$ denotes the largest and smallest singular value.}:
\begin{equation*}
\operatorname{cond}((\mathbf{U}^\top\mathbf{U}+\lambda \mathbf{I}_r)^{-1})=\frac{\sigma_1^2(\mathbf{U})+\lambda}{\sigma_n^2(\mathbf{U})+\lambda}\leq 1+\frac{\sigma_1^2(\mathbf{U})}{\lambda}.
\end{equation*}
One can see that, a proper $\lambda$ can make the output of the inversion layer stable even if the input matrix $\mathbf{U}$ is degenerated. Meanwhile, the spectral norm of $(\mathbf{U}^\top\mathbf{U}+\lambda \mathbf{I}_r)^{-1}$ satisfies $\|(\mathbf{U}^\top\mathbf{U}+\lambda \mathbf{I}_r)^{-1}\|_2\leq 1/\lambda$, which enables our inversion layer to regularize the scale of input and output factor matrices of SymNMF-Net explicitly. Furthermore, we make the inversion layer learnable. Moreover, we analyze the lower bound for $\lambda$~(illustrated in Section.~\ref{analysis}) in order to make the network converge steadily.
\par
Finally, we replace the project operation $\operatorname{max}\{\cdot,\cdot\}$ onto the nonnegative orthant with a Rectified Linear Unit (ReLU) layer and parameterize the rest of the factors $(\mathbf{X}+\lambda\mathbf{I}_n)\mathbf{U}_{k-1}$ as linear layers with weights $\mathbf{P}_k$ respectively. In summary, the construction of the SymNMF-Block is shown in Fig.~\ref{fig:SymNMF-Net-block}. As one can see, the algorithm only involves matrix inversion, multiplication and addition, which means that the forward propagation for a SymNMF-Block is only $O(nr^2)$. The details of the forward and backward propagation for our SymNet-Net are listed in the supplementary.
\begin{algorithm}[H]
	\small
	\begin{algorithmic}[1]
		\REQUIRE Input a random guess $\bU_0$ for the factorization, the data matrix $\bX\in\bbR^{n\times n}$, the weights of the SymNMF-Net $\{\bP_i\in\bbR^{n\times r}\}$, $\lambda$ and layer $K$.
		\STATE {\textbf{for} $k=1$ to $K$ \textbf{do}} 
		\STATE $\quad$ Initialize $\bP_i = (\bX+\lambda \bI_n)\bU_{i-1}$.
		\STATE $\quad$ Compute $\mathbf{U}_i = (\bX+\lambda \bI_n)\bU_{i-1}(\bU_{i-1}^T\bU_{i-1}+\lambda\bI_r)^{-1}$.
		\STATE $\quad$ Compute $\bU_i = \max(\bU_i,0).$
		\STATE  {\textbf{end for}}
		\ENSURE  The initialization of the SymNMF-Net $\{\bP_i\}$.
	\end{algorithmic}
	\caption{Initialization of the SymNMF-Net $\{\bP_i\}$.}
	\label{alg:init}
\end{algorithm}
\newpage

Just as other differential architectures, parameters $\mathbf{P}_k$ needs to be not far from the original parameter in (\ref{tradi_update}). For this account, we use classical algorithms to initial the linear layers shown in Alg.~\ref{alg:init}. By stacking the SymNMF blocks~(Fig.~\ref{fig:SymNMF-Net-block}) and then training the parameter $\{\bP_i\}$ and $\lambda$ using gradient descent method with the regression loss~($\|\bX-\bU_{out}\bU_{out}^\top\|_F$), the network can finally output an accurate factorization $\bU_{out}$ for the data matrix~($\bX$) with a random input $\bU_0$.
\par
Comparing with the state-of-the-art algorithms of problem~(\ref{prob:drop-snmf}), our method can directly optimize the original objective function of the NMF problem instead of its relaxed form, which may lead to better results. Since a pair of solutions of problem (\ref{prob:snmf}) are not always the solutions for problem (\ref{prob:drop-snmf}), and SymNMF is a non-convex problem, the search space of the local optimum sets for SymNMF-Net is much larger than those of traditional algorithms~(e.g., SymANLS and SymHALS), which means that we may find better solutions than them on different tasks. 
\par
Our method can not only be considered as a data-driven algorithm, but also can be referred as a task-driven method. If the prior knowledge can be formulated on the target matrix $\mathbf{U}$, we can augment the original loss function with corresponding regularizers and use the back-propagation algorithm to optimize the network to obtain the outputs with demanding attributes. For example, if we want to obtain a sparse approximation $\mathbf{U}$ for the original matrix $\mathbf{X}$, we can add an $\ell_1$ regularizer to the loss function as illustrated in the following experiments.
\section{Theoretical Analysis}
\label{analysis}
Since the output may greatly impact the magnitude of the gradient and others during training, the stability of the network's output plays an important role on DNN's convergence. As a result, the inversion layer is the key to the performance of our SymNMF-Net. Therefore we need to find the lower bound for $\lambda$ to make the output steady because $\lambda$ in the inversion layer matters the output a lot. 
However, we do not know the exact condition number that can ensure the stability of the output. On this account, we give two indirect conditions to maintain the stability of the output, and further use them to prove the bound which $\lambda$ needs to satisfy.
\begin{enumerate}
	\item \textbf{Proximality Condition:} Blocks of the neural network should be proximal transformations for the classical algorithm. In other words, the output of each layer needs to be not far from the output of the traditional update scheme~(\ref{tradi_update}).  
	\item \textbf{Sufficiency Condition:} The critical points of the traditional algorithm are also the critical points of the neural networks. 
\end{enumerate}
Xie et al.~\citeyearpar{dladmm} demonstrate that when the learnable parameters linear the classical optimization algorithm, from which the neural network is unfolded, the performance of the network will be better. Empirical and theoretical analyzes are given in their paper to confirm this assumption. Following the idea above, we bound the linear parameter $\mathbf{P}_i$ with the following equations:
\begin{equation}
\label{eq:bound}
\begin{aligned}
\frac{\|\mathbf{P}_i-(\mathbf{X}+\lambda \mathbf{I}_n)\tilde{\mathbf{U}}_{i-1}\|_F}{\|(\mathbf{X}+\lambda \mathbf{I}_n)\tilde{\mathbf{U}}_{i-1}\|_F}\leq \delta.
\end{aligned}
\end{equation}
In our following analysis and experiments, we call \textbf{the learnable parameter $\mathbf{P}_i$ is $\delta$-bounded} if the equations~(\ref{eq:bound}) are satisfied.
\par
In the remainder of this section, the lower bound of $\lambda$ is provided in order to achieve the above conditions.

\subsection{Notations}
Before our analysis, we list some important notations. First, we summarize the classical scheme~\ref{tradi_update} as iterating the following function:
\begin{equation*}
\tilde{\bU}_i = (\bX+\lambda\bI_n)\tilde{\bU}_{i-1}(\tilde{\bU}_{i-1}^T\tilde{\bU}_{i-1}+\lambda \bI_r)^{-1}
\end{equation*}
And we use $\tilde{\mathbf{U}}_i$ to represent the output of the classical algorithm. Meanwhile, the outputs of corresponding SymNMF-Net block are denoted as $\mathbf{U}_i$. Then we use $\mathcal{T}_U: \tilde{\mathbf{U}}_i\rightarrow\tilde{\mathbf{U}}_{i+1}$ to represent the mappings of the classical algorithm for the $i$-th iteration. We use $\mathcal{F}_U$ to denote all possible mappings of SymNMF-Net layer~(or the inference step), which can evolve to correspondingly $\mathcal{T}_U$, respectively. Finally, following the former settings, we use $\mathbf{X}$ to denote the input matrix, and let $\mathbf{P}_i$ and $\lambda$ to be learnable parameters for the $i$-th SymNMF-Net block.
\subsection{Proximality Condition}

\begin{definition}
	For any fixed $\tilde{\mathbf{U}}_i$, we call that the $i$-th block of SymNMF-Net is $\gamma$-proximal with respect to the classical algorithm $\mathcal{T}$ if the following equations hold:
	\begin{equation*}
	\begin{aligned}
	\sup_{\mathbf{U}_i\in\mathcal{B}(\tilde{\mathbf{U}}_i, \epsilon)} \|\mathcal{F}_U(\mathbf{U}_i)-\mathcal{T}_U(\tilde{\mathbf{U}}_i)\|_F&\leq C\gamma,\\
	\end{aligned}
	\end{equation*}
	where $C$ is a constant, and $\mathcal{B}(E, \gamma)=\{D | \|D-E\|_2 \leq \gamma \}$.
\end{definition}
In the following, we will prove that with certain lower bounds for $\lambda$, the block of SymNMF-Net satisfies the Proximality Condition. 
\begin{theorem}
	\label{the_1}
	Suppose that the input matrix $\|\mathbf{X}\|_2=B$, $\max_i\{\|\|\tilde{\mathbf{U}}_i\|_F, \|\mathbf{U}_i\|_F\} \leq a$, and the learnable parameter $\mathbf{P}_i$ of each block are $\epsilon$-bounded. If $\lambda$ obeys the following eqaution:
	\begin{equation*}
	\lambda>a^2+4a\epsilon,
	\end{equation*} 
	then the block is $\epsilon$-proximal with the following $C$:
	\begin{equation*}
	C =\frac{4(B+\lambda)a^2}{(\lambda-a^2)^2}+\frac{(B+\lambda)a}{\lambda-a^2}.
	\end{equation*} 
\end{theorem}
Due to limited space, for the proof of this Theorem, please refer to our Supplementary Materials. Note that since we only need to make the output stable, $\epsilon$ here need not to be too small. Moreover, empirical observations show that a small $\epsilon$ is not good for training the network, because the network will perform almost equivalent to the original update scheme~(\ref{tradi_update}). 

\subsection{Sufficiency Conditions}
The objective function for the traditional algorithm is:
\begin{equation}
\label{eq:f1}
\begin{aligned}
f_1(\mathbf{W}, \mathbf{V}) = &\frac{1}{2}\|\mathbf{X-WV}^\top\|_F^2+\frac{\lambda}{2}\|\mathbf{W-V}\|_F^2\\
&+\delta_+(\mathbf{W})+\delta_+(\mathbf{V}),
\end{aligned}
\end{equation}
where $\delta_+(\cdot)$ denotes a delta function in order to penalize the negative elements. Here, $\delta_+(x)$ is equal to $0$ if $x\geq 0$, and $+\infty$ otherwise.
Eq.~(\ref{eq:f1}) is the relaxed objective function for SymNMF. But in our network, we use the original objective function as the loss function:
\begin{equation}
\label{eq:f2}
f_2(\mathbf{U}) = \frac{1}{4}\|\mathbf{X-UU}^\top\|_F^2 +\delta_+(\mathbf{U}).
\end{equation}
The minimizers of Eq.~(\ref{eq:f2}) are not always the solutions of Eq.~(\ref{eq:f1}) for different $\lambda$. Fortunately, Zhu et al.~\citeyearpar{zhu2018dropping} give a lemma to prove that the solutions of the algorithms for problem~(\ref{prob:drop-snmf}) are the solutions of ours with proper $\lambda$.
Combining with Theorem~\ref{the_1}, we can obtain the following theorem to ensure the proximality and sufficiency conditions simultaneously.
\begin{proposition}
	\label{the_2}
	Suppose that the input matrix $\|\mathbf{X}\|_2=B$,
	$\max_i\left\{\|\tilde{\mathbf{U}}_i\|_F, \|\mathbf{U}_i\|_F\right\} \leq a$, and the parameters $\mathbf{P}_i$ of SymNMF-Net are $\epsilon$-bounded. If $\lambda$ satisfies the following inequality:
	\begin{equation*}
	\lambda > \max\{a^2+4a\epsilon, \frac{1}{2}(\|\mathbf{X}\|_F+\|\mathbf{X}-\mathbf{U}_0\mathbf{U}_0^\top\|_F)\}
	\end{equation*}
	then SymNMF-Net satisfies the proximality and sufficiency conditions simultaneously.
	
\end{proposition}
In brief, if $\lambda$ stays in the above region stated in Prop.~\ref{the_2}, the stability of the SymNMF-Net's output is ensured. With the theoretical results of the perturbed gradient descent method~\cite{jin2017escape}, our network will finally converge to the local minima of the SymNMF problem.

\section{Numerical Experiments}
\label{exp-sec}
In this section, we conduct empirical results to verify the numerical convergence of SymNMF-Net, the sufficiency of lower bound for $\lambda$, and the superior performance of the factorization on image and text clustering tasks.
\par
SymNMF can be used for graph clustering by factorizing the similarity matrix $\mathbf{X}$ of the data~\citep{kuang2015symnmf}. In this part, we use the graph clustering methods on different image datasets and text datasets to evaluate the performance of SymANLS, SymHALS and SymNMF-Net.

\paragraph{Loss Functions:} According to the conclusions by Xie et al.~\citeyearpar{dladmm} that the learnable parameters would better lie near the original optimization algorithm, we add regularizers for $\mathbf{P}$ to the loss function:  
\begin{equation*}
\begin{aligned}
\text{Loss} =& \sum_{i=1}^L(\|\mathbf{X}-\bU_i\bU_i^\top\|_F^2 +\beta\|\mathbf{P}_i-(\mathbf{X}+\lambda \mathbf{I}_n)\mathbf{U}_{i-1}\|_F^2),
\end{aligned}
\end{equation*}
where $L$ represents the layer number of SymNMF-Net.

\paragraph{Training Details:} Different from SymANLS and SymHALS, SymNMF-Net has learnable parameters and can use Back-Propagation for training the whole network. We stack the network with $10$ SymNMF blocks and use a GTX$-1080$Ti GPU for our experiments below. Adam~\citep{kingma2014adam} is adopted to update the parameters with the learning rate starting from $lr=0.5$ for image clustering and $lr=0.3$ for text clustering. We then let $\beta=5e-6$, and let the learnable parameter $\lambda$ initialize to $\|\mathbf{X}\|_F$.\footnote{The lower bound of $\lambda$ is around $\|\mathbf{X}\|_F$ when the training process starts.}
\subsection{Numerical Analysis}

\paragraph{Convergence Evaluation}
Firstly, we conduct experiments on ORL and COIL datasets to compare the numerical error among SymANLS, SymHALS, and our SymNMF-Net. We evaluate the performance with the relative error of the output~$\mathbf{U}$ and the input matrix $\mathbf{X}$:
\begin{equation*}
\text{(Relative Error)\quad}E = \frac{\|\mathbf{X}-\mathbf{U}\mathbf{U}^\top\|_F^2}{\|\mathbf{X}\|_F^2}.
\end{equation*}

\begin{figure}[t]
	\begin{minipage}[t]{0.21\textwidth}
		\centering
		\includegraphics[width=1.17\textwidth]{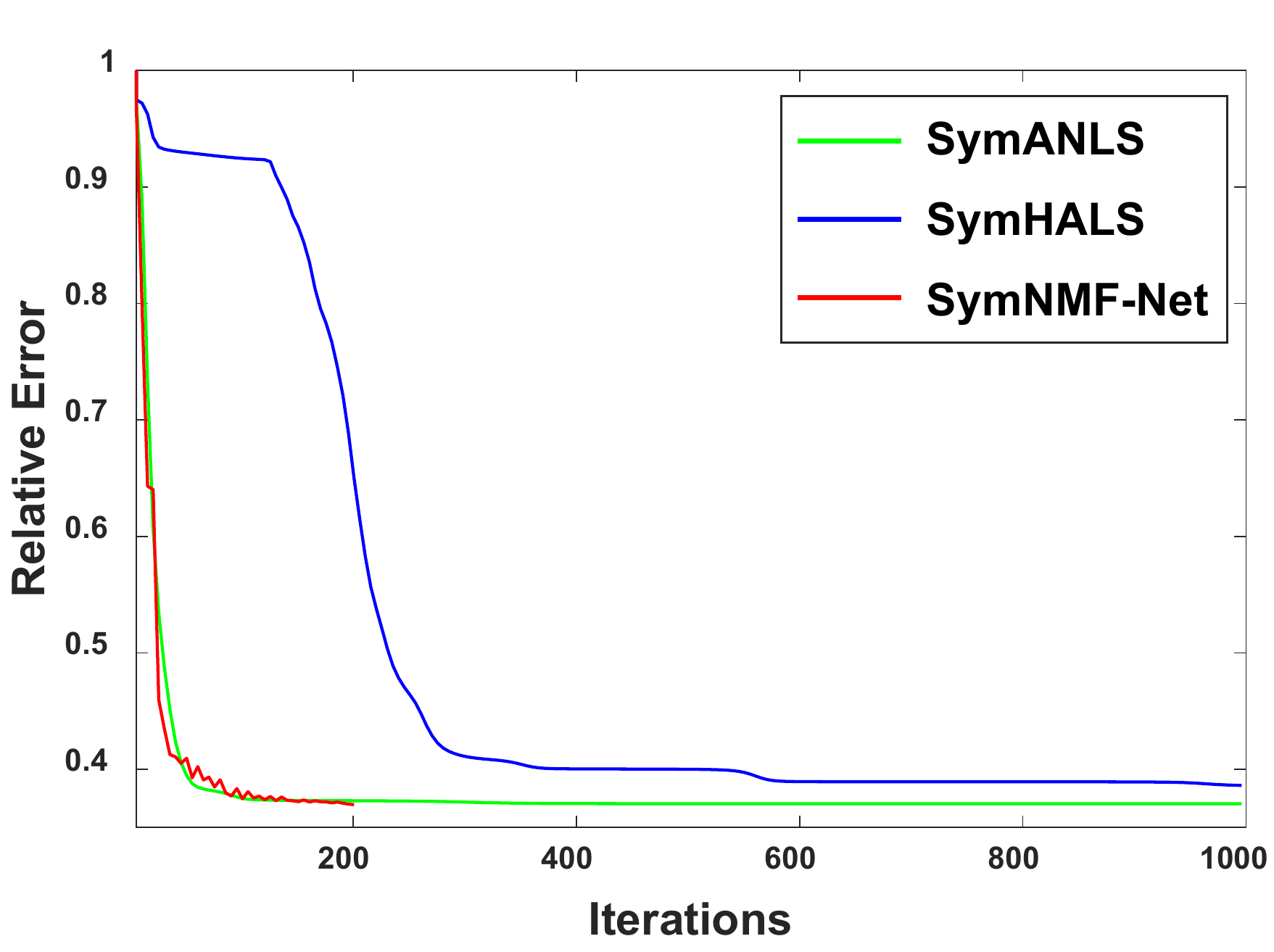}
		\centerline{\qquad(a)}
	\end{minipage}
	\qquad
	\begin{minipage}[t]{0.21\textwidth}
		\centering
		\includegraphics[width=1.2\textwidth]{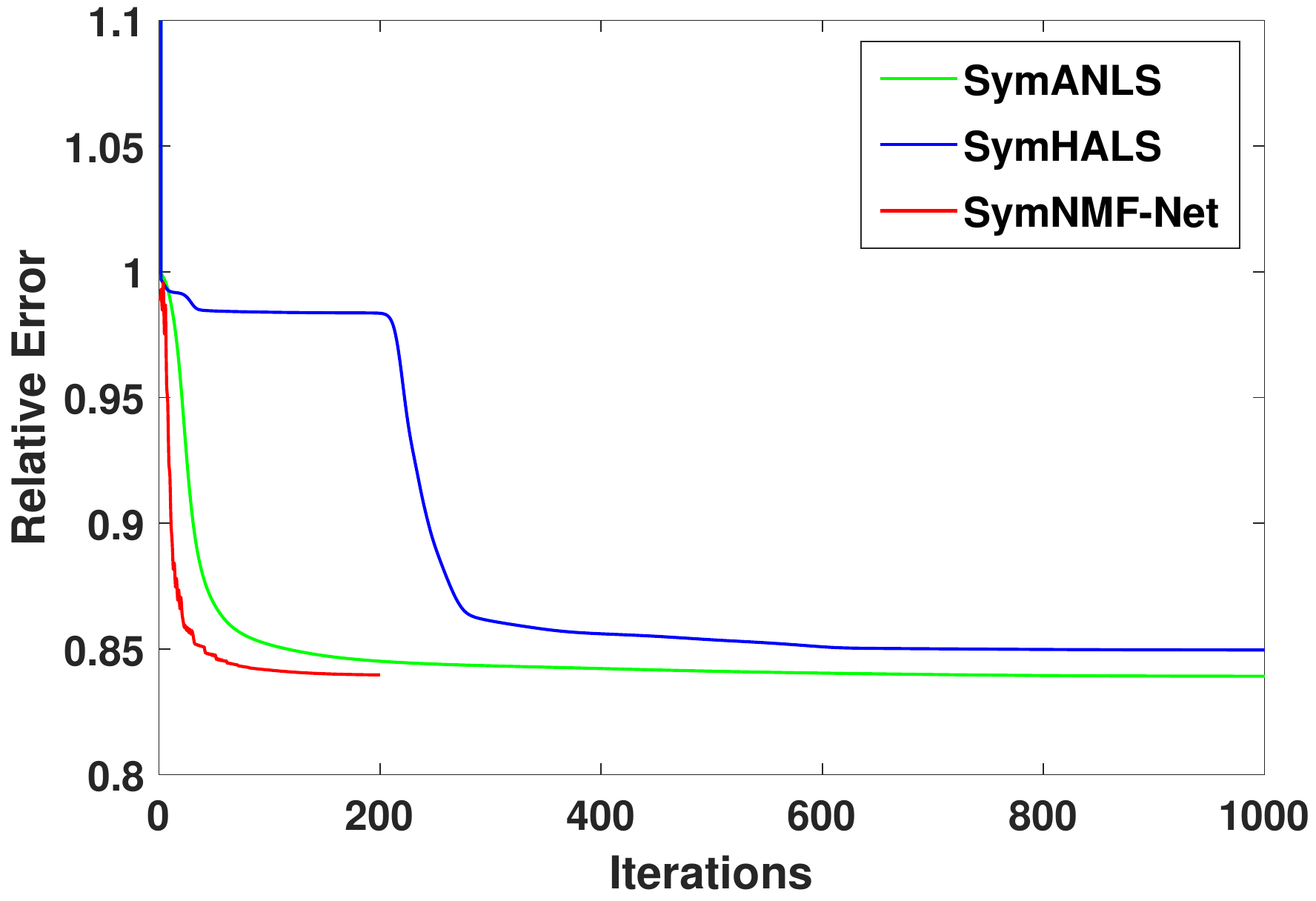}
		\centerline{\qquad(b)}
	\end{minipage}
	\caption{The numerical relative error of SymANLS, SymHALS and SymNMF-Net on (a) ORL and (b) COIL-$20$.}
	\label{fig:convergence}
\end{figure}

We use the similarity matrix of COIL-$20$ and ORL dataset to compare the convergence property of different models~(shown in Figure~\ref{fig:convergence}). Our SymNMF-Net achieves superior performance than SymHALS and the convergence speed is comparable with SymANLS on ORL dataset. Meanwhile, our model converges much faster than both SymANLS and SymHALS on COIL-$20$. Furthermore, the relative error of our model is better than others~(Table.~\ref{tab:relative-error}). In general, the empirical results confirmed that our model can achieve better performance than classical algorithms under numerical evaluation. 
\begin{table}[t]
	\centering
	\setlength{\tabcolsep}{1.0mm}{
		\begin{tabular}{|c|c|c|}
			\hline
			& ORL      & COIL-20 \\
			\hline
			SymANLS     &  $0.2669$ & $0.8405$\\
			\hline
			SymHALS     &  $0.2723$ & $0.8501$\\
			\hline
			\textbf{SymNMF-Net} &  $\mathbf{0.2659}$ & $\mathbf{0.8390}$\\
			\hline
		\end{tabular}
		\caption{The final relative error for different models on ORL and COIL-$20$.}
		\vspace{-3mm}
		\label{tab:relative-error}
	}
	\label{tab:relative_error}
\end{table}
\begin{figure}[h]
	\centering
	\includegraphics[width=0.4\textwidth]{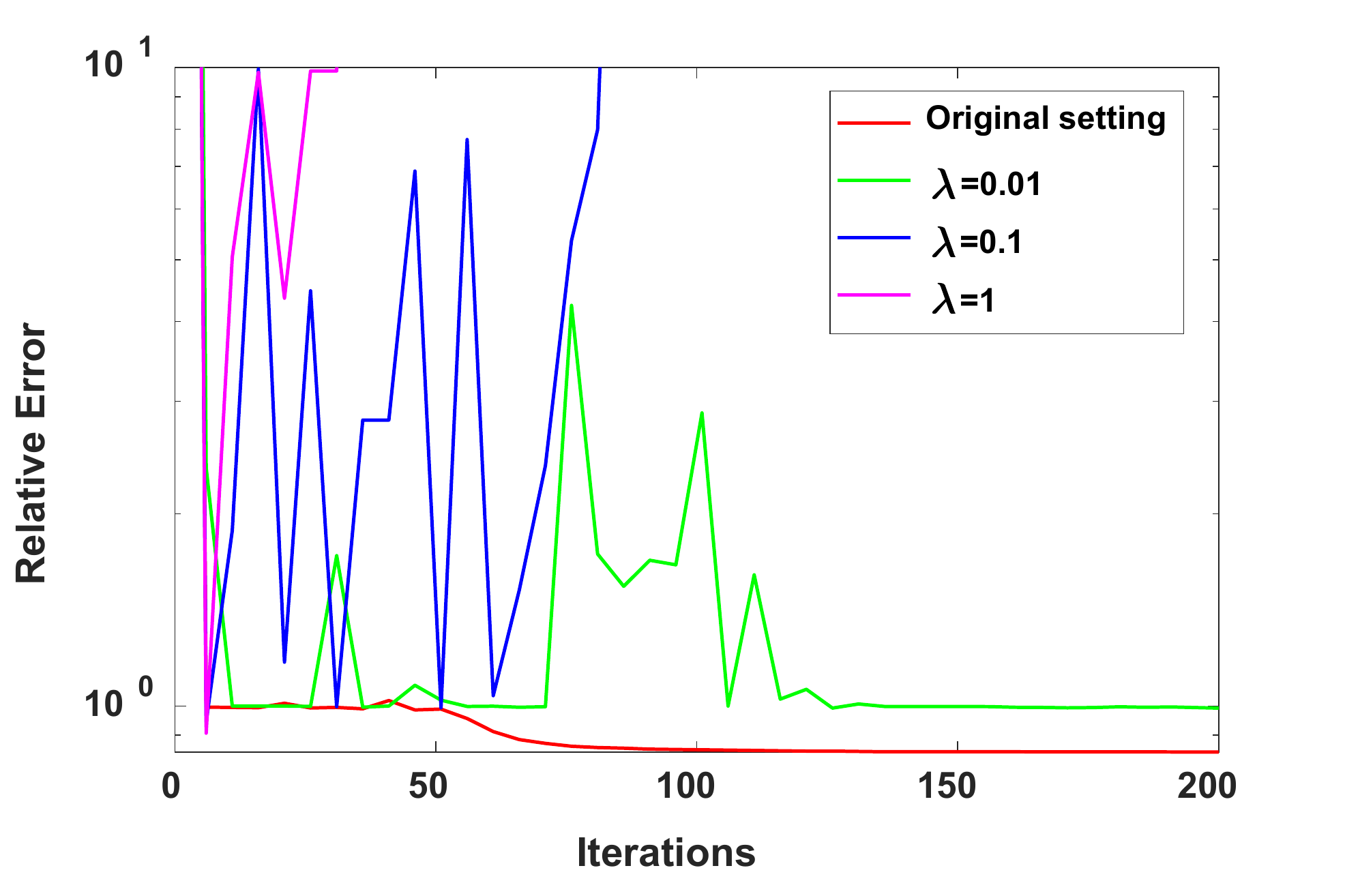}
	\caption{SymNMF-Net with different $\lambda$ on COIL-$20$ dataset.}
	\label{fig:lam}
\end{figure}
\paragraph{Sparse output with $\ell_1$ regularizer}
Since the SymNMF problem is a non-convex problem, there are a variety of local minima with different characteristics like sparsity. Nevertheless, classical algorithms are hard to involve certain constraints. Consequently, they are difficult to obtain results with demanding attributes. In contrast, our SymNMF-Net can obtain different results with demanding properties by modifying the loss functions.
\begin{figure}[h]
	\centering
	\includegraphics[width=0.35\textwidth]{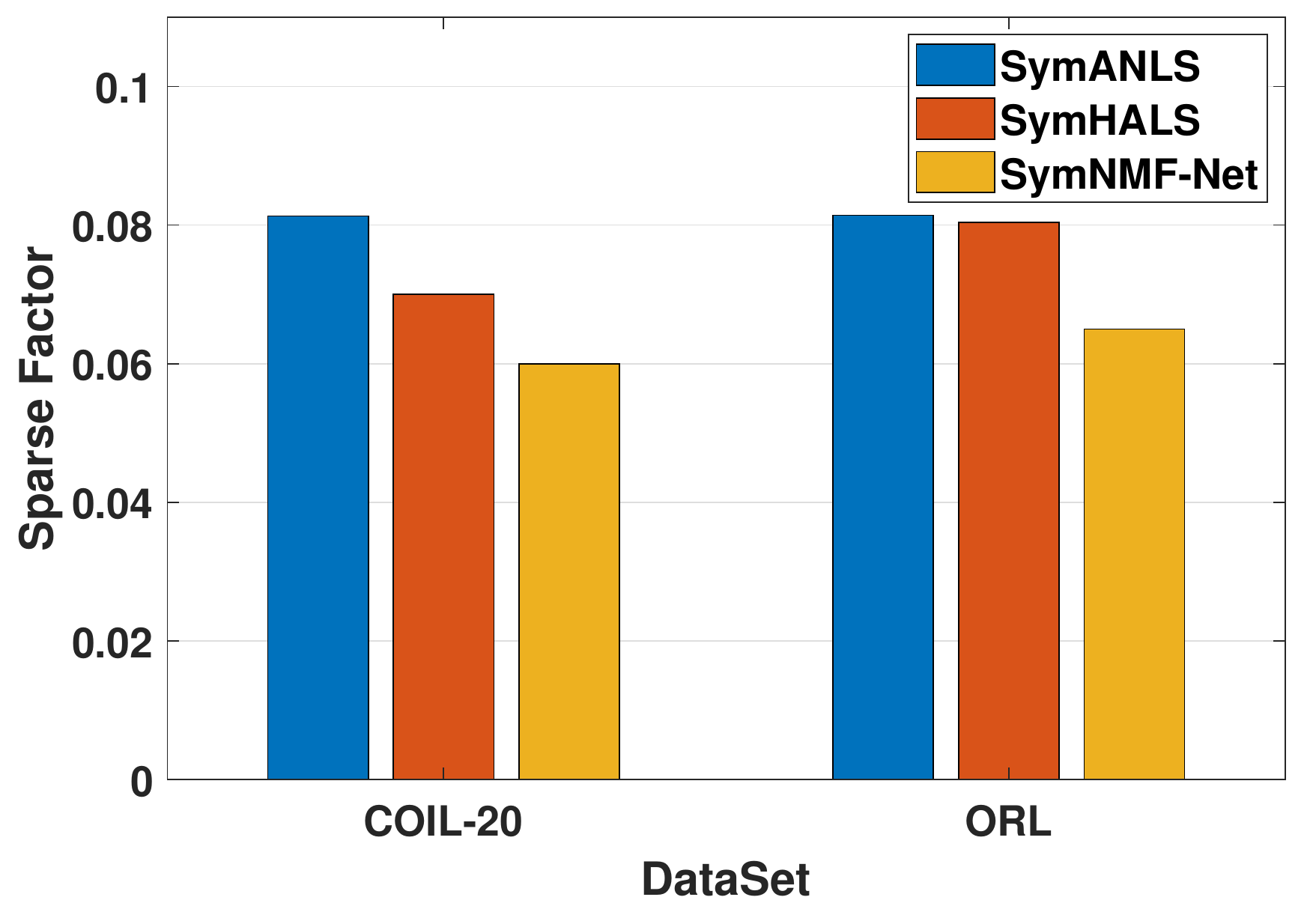}
	\caption{The Sparse Factor~(SF) of the output factor matrix.}
	\label{fig:L}
	\vspace{-0.5cm}
\end{figure}

\par
For instance, our SymNMF-Net can generate more sparse factor matrix than the classical products by adding $\ell_1$ norm of the outputs to the objective function. We compare our model with the state-of-the-art algorithms on COIL-$20$, ORL and evaluate the sparse factor~(CF) of the outputs, which is a widely used indicator to measure the sparsity. The formulation of SF is illustrated as follow:
\begin{equation*}
SF(x):=\frac{\#\{i,|x_i|>T\}}{\#\{\bx\}}
\end{equation*}
where $T$ is the sparse threshold which we set to be $0.01*mean(\bx)$. All the three methods produce the results with similar relative error but different in sparseness shown in Figure.~\ref{fig:L}. As we expected, the decomposition from our model is the most sparse one among the three methods, which claimed our motivation for constructing an architecture which can handle non-negative matrix factorization with different expectations on the output.
\paragraph{The effect of $\lambda$} To study the effects of the parameter $\lambda$ and validate the analyses in Section~\ref{analysis}, we show the relative error versus iteration for different $\lambda$ on COIL-$20$ in Figure~\ref{fig:lam}. The original setting is that we initialize the learnable parameter $\lambda$ with $\|\mathbf{X}\|_F~(\approx 11.2)$, and ensure it obeys our lower bound during the training process by a projection. For the other cases, we fix the $\lambda$ as listed in the legends in the course of the training process. As we can see from Figure~\ref{fig:lam}, the network does not converge if $\lambda$ is small, which validates the sufficiency of the conditions in Theorem~\ref{the_2} that $\lambda$ should have a lower bound.

\subsection{Image Clustering}
In the following two sections, we use the SymNMF based graph clustering methods on datasets to evaluate the performance of the numerical algorithms and our SymNMF-Net.

\paragraph{SymNMF for Clustering:}

Putting all the images or texts to be clustered into a data matrix $\mathbf{M}$ whose rows are vactorized images or texts, we can construct the similarity matrix $\mathbf{X}$ following the procedures in previous works~\citep{kuang2015symnmf,zhu2018dropping}. After running our SymNMF-Net or traditional algorithms on the similarity matrix $\mathbf{X}$, the non-negative approximation $\mathbf{U}$ can be obtained, then the label of the $i$-th sample can be obtained by:
\begin{equation*}
l(\mathbf{M}_i) = \mathop{\arg\max}_j \tilde{\mathbf{U}}_{ij}.
\end{equation*}
The experiments are conducted on the following image datasets for face, instances or digits:
\begin{table}[t]
	\small
	\centering
	\setlength{\tabcolsep}{1.0mm}{
		\begin{tabular}{|c|c|c|c|c|c|}
			\hline
			& ORL      & COIL-20 & MNIST$_{1}$ & MNIST$_{2}$ \\
			\hline
			SymANLS     &  $80.75$ & $79.79$ & $64.77$ & $85.89$\\
			\hline
			SymHALS     &  $75.70$ & $58.54$ & $66.57$ & $86.08$\\
			\hline
			ADMM        &  $76.50$ & $69.03$ & $58.03$ & $87.13$\\
			\hline
			SymNewton   &  $76.25$ & $74.72$ & $59.90$ & $85.89$\\
			\hline
			PGD         &  $77.00$ & $72.43$ & $64.75$ & $87.10$\\
			\hline
			\textbf{SymNMF-Net} &  $\mathbf{81.50}$ & $\mathbf{80.13}$ &$\mathbf{70.64}$  & $\mathbf{98.98}$\\
			\hline
		\end{tabular}
		\caption{Summary of image clustering accuracy~(\%) for different algorithms on different image datasets.}
		\label{tab:imgclustersum}}
	\end{table}

\paragraph{ORL:} The dataset has 400 gray-scale images for $40$ distinct persons, the size of which is $92\times 112$. Each person has ten different photos taken from different angles and emotions.\footnote{http://www.cl.cam.ac.uk/research/dtg/attarchive/face-database.html\ \ }

\paragraph{COIL-20:} It consists of $1,440$ images of $20$ objects. Each instance is a $128\times 128$ gray-scale image.\footnote{http://www.cs.columbia.edu/CAVE/software/softlib/coil-20.php}

\paragraph{MNIST:} MNIST is a classical handwritten digits dataset containing $60,000$ $32\times32$ gray-scale images for training and $10,000$ images for testing. Following the setting of the previous work~\citep{zhu2018dropping}, we use $3,147$ images in the training set for $10$ classes (denoted as MNIST$_{1}$) and $3,147$ images in the test set which belong to $3$ digits (denoted as MNIST$_{2}$).\footnote{http://yann.lecun.com/exdb/mnist/}
\par
Comparing the clustering accuracy~(shown in Table~\ref{tab:imgclustersum} and Table~\ref{tab:orl}) we can see that, our SymNMF-Net performs much better than all other state-of-the-art graph clustering methods on different evaluation methods for clustering. 
\begin{table}[t]
	\centering
	\begin{tabular}{|c|c|c|c|c|}
		\hline
		&  \multicolumn{2}{c|}{ORL} & \multicolumn{2}{c|}{COIL-$20$}  \\
		\hline
		&  NMI &  Purity  &  NMI &  Purity\\
		\hline
		SymANLS   &  $0.884$  & $0.805$ &  $0.849$ & $0.804$ \\
		\hline 
		SymHALS   &  $0.862$  & $0.763$ &  $0.664$ & $0.538$\\
		\hline
		\textbf{SymNMF-Net}   &  $\mathbf{0.904}$  & $\mathbf{0.835}$ &  $\mathbf{0.896}$  & $\mathbf{0.853}$ \\
		\hline
	\end{tabular}
	\caption{Summary of image clustering Purity and Normalized Mutual Information~(NMI) for SymNMF-Net~(ours), SymANLS and SymHALS on ORL and COIL.}
	\label{tab:orl}
\end{table}
\par
\begin{figure}[t]
	\begin{minipage}[t]{0.21\textwidth}
		\centering
		\includegraphics[width=1.19\textwidth]{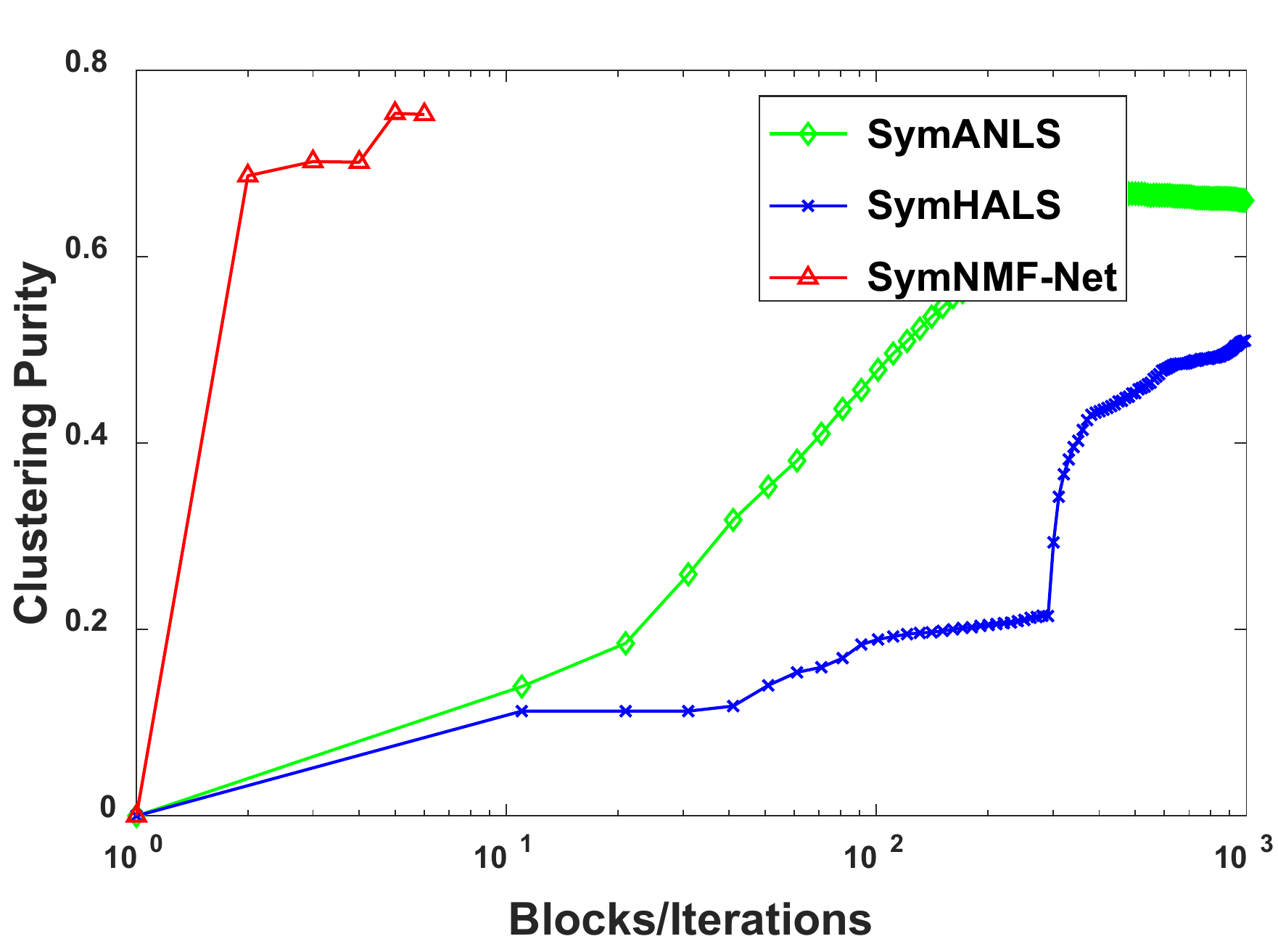}
		\centerline{\qquad(a)}
	\end{minipage}
	\qquad
	\begin{minipage}[t]{0.21\textwidth}
		\centering
		\includegraphics[width=1.2\textwidth]{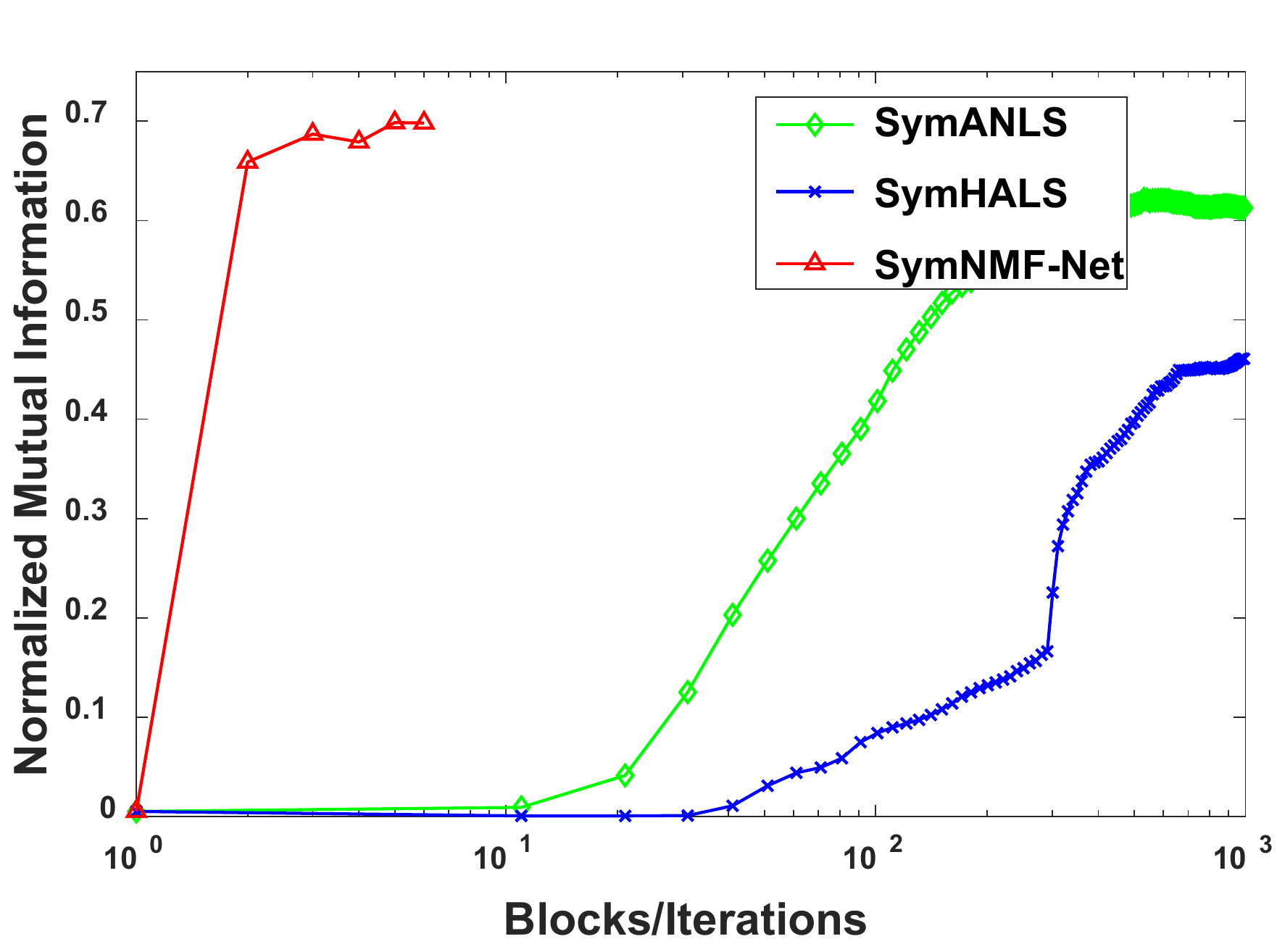}
		\centerline{\qquad(b)}
	\end{minipage}
	\caption{The curves of (a) clustering purity and (b) NMI with respect to iterations/blocks of SymNMF-Net, SymANLS and SymHALS on MNIST$_{1}$.}
	\label{fig:mnist_train}
\end{figure}
\begin{figure}[t]
	
	\begin{minipage}[t]{0.21\textwidth}
		\centering
		\includegraphics[width=1.665in]{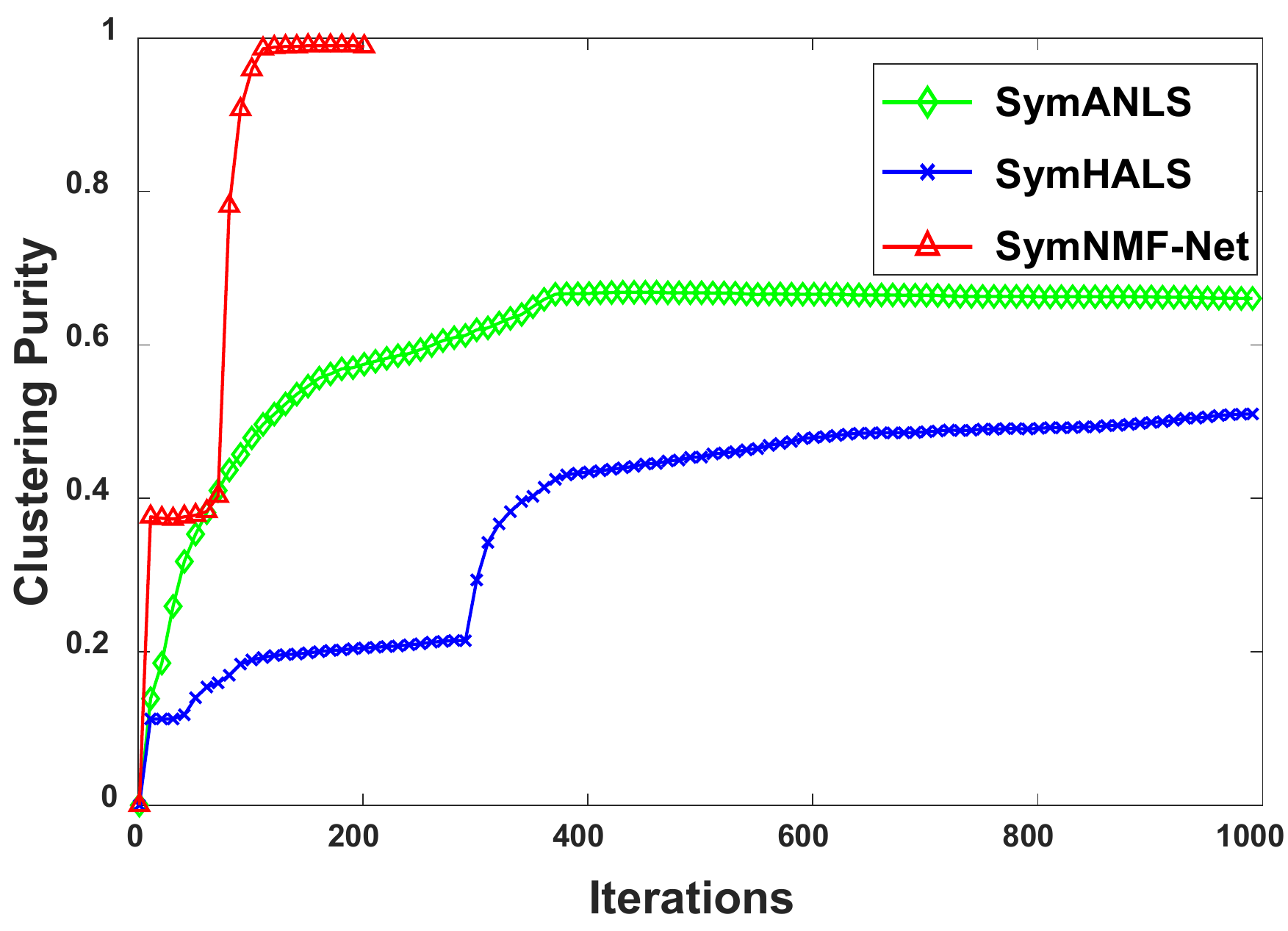}
		\centerline{\qquad(a)}
	\end{minipage}
	\qquad
	\begin{minipage}[t]{0.21\textwidth}
		\centering
		\includegraphics[width=1.68in]{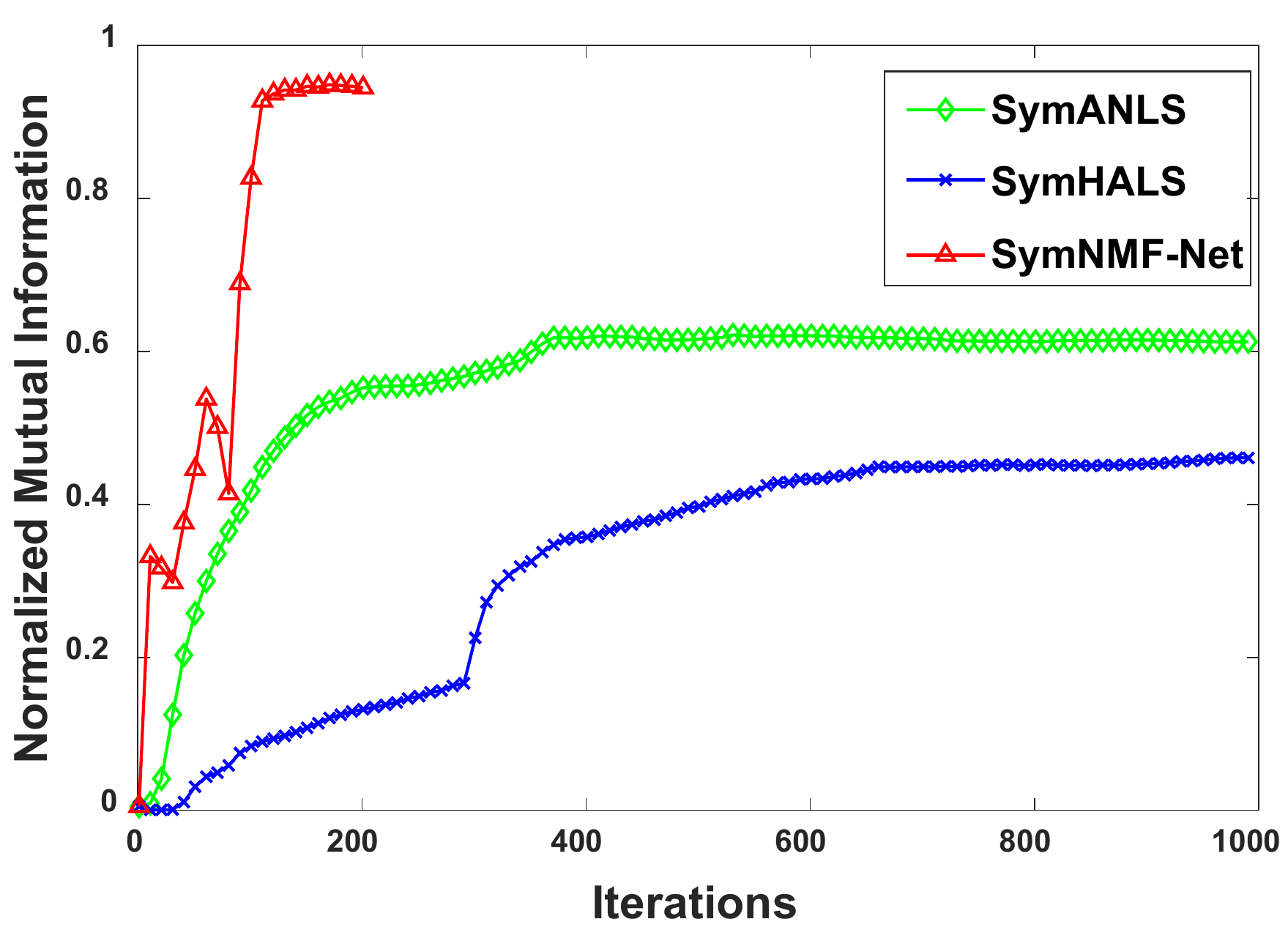}
		\centerline{\qquad(b)}
	\end{minipage}
	\caption{The curves of (a) clustering purity and (b) NMI with respect to training or running iterations of SymNMF-Net, SymANLS and SymHALS on MNIST$_{2}$.}
	\label{fig:mnist_test}
\end{figure}

Furthermore, the performances of SymNMF-Net comparing with SymHALS and SymANLS on MNIST$_{1}$ and MNIST$_{2}$ are shown in Figures~\ref{fig:mnist_train} and \ref{fig:mnist_test}, respectively. From these two figures, one can see that our SymNMF-Net beat them with a large margin on both datasets in terms of both metrics cluster purity and NMI. Note that the inference of our SymNMF-Net can be regarded as an alternating algorithm for SymNMF~(\ref{prob:drop-snmf}) with $5$ iterations, therefore we draw the curves for the clustering purity and NMI versus iterations for SymNMF-Net, SymANLS and SymHALS, which are shown in Figure~\ref{fig:mnist_train} and Figure~\ref{fig:mnist_test}. As can be seen that, the alternating algorithm represented by our network enjoys a huge advantages for speed and performance on clustering.

In addition, from the curves of clustering purity and NMI w.r.t. training iterations in Figure~\ref{fig:mnist_test} we can see that, our SymNMF-Net is still able to overperform SymANLS and SymHALS for clustering purity and NMI even with less iterations.


\subsection{Text Clustering}
In addition to image clustering , we also finish the experiments on text data.
\par
\paragraph{TDT-$2$:} The TDT-$2$ corpus contains data collected during the first half of $1998$ from newswires, radio programs and television programs.  We use its largest $30$ categories for the experiments, which contain $9,394$ documents in total. {\footnote{https://www.ldc.upenn.edu/collaborations/past-projects}} 

\begin{table}[t]
	\centering
	\begin{tabular}{|c|c|c|c|}
		\hline
		&  \multicolumn{3}{c|}{TDT$-2$}  \\
		\cline{2-4}
		&  ACC      & NMI   & Purity \\
		\hline
		SymANLS     &  $75.36$  & $0.746$ & $0.794$  \\
		\hline
		SymHALS     &  $81.31$  & $0.755$ & $0.833$  \\
		\hline
		\textbf{SymNMF-Net (ours)}       &  $\mathbf{82.34}$  & $\mathbf{0.807}$ & $\mathbf{0.859}$\\
		\hline
	\end{tabular}
	\caption{Summary of graph clustering accuracy~(\%), normalized mutual information(NMI), and clustering purity of SymNMF-Net~(ours), SymANLS, and SymHALS on TDT$-2$.}
	\label{tab:tdt}
\end{table}


As can be seen from Table~\ref{tab:tdt}, no matter using clustering accuracy, purity, or NMI as metrics for evaluation, the performance of SymNMF-Net is much better than other comparative methods on both TDT-$2$. 

Moreover, there is an important phenomenon from the two sections that, SymANLS performs better than SymHALS on image clustering whereas worse than SymHALS on text clustering. This mainly because these two methods are not stable on clustering task, meanwhile our SymNMF-Net overperforms these two methods on both text and image clustering tasks. The results of this comparison can demonstrate the motivation to a certain extent that, our SymNMF-Net is better at adapting to the tasks and the datasets than SymANLS and SymHALS. 
\section{Conclusion}
In this paper, we propose a deep-learning-based optimization method called SymNMF-Net for solving the SymNMF problem. In particular, each block of our network contains an inversion layer, a linear layer and ReLU. Moreover, we analyze the conditions which can ensure the output stability of each inversion layer. We further show that our SymNMF-Net shares some critical points with traditional methods under our conditions. 
Besides, the empirical results demonstrate that our model finally evolves to a better data-driven algorithm which converges much faster than the traditional optimization algorithms. 
Experiments also show that, comparing with other state-of-the-art SymNMF methods on graph clustering, our model consistently leads to a better performance. 
In this paper, we have proposed an inversion in neural network and designed our SymNMF-Net, a differentiable architecture for SymNMF problem. Experiments have been done to show our model will leads to consistently better performance on model convergence and clustering. Besides, the empirical results demonstrate that our model finally evolves to a better data-driven algorithm which converges much faster than traditional optimization problem. 
\newpage
\bibliographystyle{named}
\bibliography{ijcai20}
\end{document}


\maketitle
	
	\section{Forward and Backward Scheme for SymNMF-Net}
	With a random initialized guess for the factorization matrix, the forward scheme for a SymNMF-Net with $K$ blocks can be stated as follow:
	\begin{algorithm}[H]
		\small
		\begin{algorithmic}[1]
			\REQUIRE Initial guess $\bU_0$, fully-connected layers $\{FC_i\}$ with parameter $\{\bP_i\}\in\bbR^{n\times r}$, learnable parameter $\lambda\in\bbR$ and block number $K$.
			\STATE  {\textbf{for} $k=1$ to $K$ \textbf{do}}
			\STATE $\quad$ Compute $\mathbf{INV}_i = (\bU_{i-1}^T\bU_{i-1}+\lambda\bI_r)^{-1} \in\bbR^{r\times r}$ .
			\STATE $\quad$ Compute $\bU_i =  FC_i(\mathbf{INV}_i) \in\bbR^{n\times r}.$
			\STATE $\quad$ Compute $\bU_i = ReLU(\bU_i).$
			\STATE  {\textbf{end for}}
			\STATE $\bU_{out} = \bU_K$
			\ENSURE  The output of the SymNMF-Net $\bU_{out}$.
		\end{algorithmic}
		\caption{The forward propagation for the SymNMF-Net.}
		\label{alg:sym-fp}
	\end{algorithm}
	Then we use gradient descent to train the learnable parameter $\{\bP_i\}$ and $\lambda$, which are listed in the following:
	\begin{algorithm}[H]
		\small
		\begin{algorithmic}[1]
			\REQUIRE The data matrix $\bX$, loss function $\cL$, output of the SymNMF-Net $\bU_{out}$, learning rate $\alpha$, block number $K$, the learnable parameters $\{\bP_i\}$ and $\lambda$.
			\STATE  {\textbf{for} $k=K$ to $1$ \textbf{do}}
			\STATE $\quad$ Update $\mathbf{P}_i = \bP_i -\alpha\frac{\partial \cL(\bX,\bU_{out})}{\partial \bP_i}$.
			\STATE  {\textbf{end for}}
			\STATE Update $\lambda=  \frac{\partial \cL(\bX,\bU_{out})}{\partial \lambda}$.	
		\end{algorithmic}
		\caption{The $i$-th SymNMF Block.}
		\label{alg:sym-bp}
	\end{algorithm}
	
	\section{Proof for the Theorem}
	\subsection{Notations}
	Before our analysis, we list some important notations. First, we summarize the classical scheme~\ref{tradi_update} as iterating the following function:
	\begin{equation*}
	\tilde{\bU}_i = (\bX+\lambda\bI_n)\tilde{\bU}_{i-1}(\tilde{\bU}_{i-1}^T\tilde{\bU}_{i-1}+\lambda \bI_r)^{-1}
	\end{equation*}
	And we use $\tilde{\mathbf{U}}_i$ to represent the output of the classical algorithm. Meanwhile, the outputs of corresponding SymNMF-Net block are denoted as $\mathbf{U}_i$. Then we use $\mathcal{T}_U: \tilde{\mathbf{U}}_i\rightarrow\tilde{\mathbf{U}}_{i+1}$ to represent the mappings of the classical algorithm for the $i$-th iteration. We use $\mathcal{F}_U$ to denote all possible mappings of SymNMF-Net layer~(or the inference step), which can evolve to correspondingly $\mathcal{T}_U$, respectively. Finally, following the former settings, we use $\mathbf{X}$ to denote the input matrix, and let $\mathbf{P}_i$ and $\lambda$ to be learnable parameters for the $i$-th SymNMF-Net block.
	
	\subsection{Proof}
	\begin{theorem}
		\label{the_1}
		Suppose that the input matrix $\|\mathbf{X}\|_2=B$, $\max_i\{\|\|\tilde{\mathbf{U}}_i\|_F,\|\mathbf{U}_i\|_F\} \leq a$, and the learnable parameter $\mathbf{P}_i$ of each block are $\epsilon$-bounded. If $\lambda$ obeys the following eqaution:
		\begin{equation*}
		\lambda>a^2+4a\epsilon,
		\end{equation*} 
		then the block is $\epsilon$-proximal with the following $C$:
		\begin{equation*}
		C =\frac{4(B+\lambda)a^2}{(\lambda-a^2)^2}+\frac{(B+\lambda)a}{\lambda-a^2}.
		\end{equation*} 
	\end{theorem}
	Before we prove the theorem above, we some present useful results for matrix inversion and the bounds of the output.
	\begin{lemma}
		For given matrices $\mathbf{A}$ and $\mathbf{B}=\mathbf{A}+\Delta$ with $\|\mathbf{A}^{-1}\Delta\|<1$, the following equations are always hold:
		\begin{equation*}
		\begin{aligned}
		\|\mathbf{B}^{-1}-\mathbf{A}^{-1}\|\leq& \frac{\|\mathbf{A}^{-1}\|^2\|\Delta\|}{1-\|\mathbf{A}^{-1}\Delta\|}\\
		\|\mathbf{B}^{-1}\|\leq & \frac{\|\mathbf{A}^{-1}\|}{1-\|\mathbf{A}^{-1}\Delta\|}
		\end{aligned}
		\end{equation*}
		where $\|\cdot\|$ represents all matrix norms which are submultiplicative.
	\end{lemma}
	
	Using the lemma above, we can give the proof for the Theorem \ref{the_1}.
	
	\begin{proof}
		Denoting the original algorithm as an operator $\mathcal{T}_k$ for its $i-$th iteration~(we choose $i$ randomly) on matrix $\tilde{\mathbf{U}}$ before projection:
		\begin{equation*}
		\tilde{\mathcal{T}}_{Ui}(\tilde{\mathbf{U}}_i)=(\mathbf{X}+\lambda\mathbf{I}_n)\tilde{\mathbf{U}}_i(\tilde{\mathbf{U}}_i^\top\mathbf{U}_i+\lambda\mathbf{I}_r)^{-1},
		\end{equation*}
		while the neural network learns an operator $\mathcal{F}_{Ui}$ on matrix $\mathbf{U}\in\mathcal{B}(\tilde{\mathbf{U}},\epsilon)$ before ReLU:
		\begin{equation*}
		\tilde{\mathcal{F}}_{Ui}(\mathbf{U}_i)=\mathbf{P}_i(\mathbf{U}_i^\top\mathbf{U}_i+\lambda\mathbf{I}_r)^{-1},
		\end{equation*}
		then, the operator for original algorithm with projection and neural network with ReLU is defined as $\mathcal{T}=ReLU(\tilde{\mathcal{T}})$ and $\mathcal{F}=ReLU(\tilde{\mathcal{F}})$.
		\par
		Secondly, because $\max_i\{\|\|\tilde{\mathbf{U}}_i\|_F,\|\mathbf{U}_i\|_F\} \leq a$, then the following equations hold:
		\begin{equation*}
		\max_i\{\|\|\tilde{\mathbf{U}}_i\|_2,\|\mathbf{U}_i\|_2\} \leq a.
		\end{equation*}
		Then from the second equation of the above lemma, for every $\mathbf{U_i}$ we can get:
		\begin{equation*}
		\begin{aligned}
		\|(\mathbf{U}_i\mathbf{U}_i^\top+\lambda\mathbf{I}_r)^{-1}\|_2 
		&\leq\frac{\|(\lambda \mathbf{I}_r)^{-1}\|_2}{1-\|(\lambda \mathbf{I})^{-1}\mathbf{U}_i\mathbf{U}_i^\top\|_2}\\
		&=\frac{\frac{1}{\lambda}}{1-\frac{1}{\lambda}\|\mathbf{U}_i\mathbf{U}_i^\top\|_2}\\
		&\leq \frac{1}{\lambda-a^2},
		\end{aligned}
		\end{equation*}
		And using the triangle inequality, the following equations:
		\begin{equation*}
		\begin{aligned}
		&\|\mathbf{U}_i^\top\mathbf{U}_i-\tilde{\mathbf{U_i}}^\top\tilde{\mathbf{U_i}}\|_2\\
		\leq&\|\mathbf{U}_i^\top(\mathbf{U}_i-\tilde{\mathbf{U}}_i)+(\mathbf{U_i}-\tilde{\mathbf{U}}_i)^\top\tilde{\mathbf{U}}_i\|_2\\
		\leq&\|\mathbf{U}_i\|_2\|\mathbf{U}_i-\tilde{\mathbf{U}}_i\|_2+\|\tilde{\mathbf{U}}_i\|_2\|\mathbf{U}_i-\tilde{\mathbf{U}}_i\|_2\\
		\leq& 2a\epsilon.
		\end{aligned}
		\end{equation*}
		With the first equation of the above lemma, $\forall \mathbf{U}_i\in\mathcal{B}(\tilde{\mathbf{U}}_i,\epsilon)$, we can get the following equations since $\lambda>a^2+4a\epsilon$,
		\begin{equation*}
		\begin{aligned}
		&\|(\tilde{\mathbf{U}}_i^\top\tilde{\mathbf{U}}_i+\lambda\mathbf{I}_r)^{-1}-(\mathbf{U}_i^\top\mathbf{U}_i+\lambda\mathbf{I}_r)^{-1}\|_2\\
		\leq&\frac{\|(\mathbf{U}_i^\top\mathbf{U}_i+\lambda\mathbf{I}_r)^{-1}\|_2^2\|\mathbf{U}_i^\top\mathbf{U}-\tilde{\mathbf{U_i}}^\top\tilde{\mathbf{U_i}}\|_2}{1-\|(\mathbf{U}_i^\top\mathbf{U}_i+\lambda\mathbf{I}_r)^{-1}\|_2\|\mathbf{U}_i^\top\mathbf{U}-\tilde{\mathbf{U_i}}^\top\tilde{\mathbf{U_i}}\|_2}\\
		\leq & \frac{\frac{1}{(\lambda-a^2)^2}2a\epsilon}{1-\frac{2a\epsilon}{\lambda-a^2}} 
		=  \frac{2a}{(\lambda-a^2)(\lambda-a^2-2a\epsilon)}\epsilon.
		\end{aligned}
		\end{equation*}
		Since $\mathbf{P}_i$ is $\epsilon$-bounded, which means:
		\begin{equation*}
			\|\mathbf{P}_i-(\mathbf{X}+\lambda\mathbf{I}_n)\tilde{\mathbf{U}}_i\|_F \leq  \|(\mathbf{X}+\lambda\mathbf{I}_n)\|_2\|\tilde{\mathbf{U}}_i\|_F\epsilon.
		\end{equation*}
		In this way, we can get:
		\begin{equation*}
		\begin{aligned}
		&\|\tilde{\mathcal{F}}_{Ui}(\mathbf{U}_i)-\tilde{\mathcal{T}}_{Ui}(\tilde{\mathbf{U}}_i)\|_F\\
		\leq&\|(\mathbf{X}+\lambda\mathbf{I}_n)\tilde{\mathbf{U}}_i\|_F\|(\tilde{\mathbf{U}}_i^\top\tilde{\mathbf{U}}_i+\lambda\mathbf{I}_r)^{-1}-(\mathbf{U}_i^     T\mathbf{U}_i+\lambda\mathbf{I}_r)^{-1}\|_2\\
		& + \|\mathbf{P}_i-(\mathbf{X}+\lambda\mathbf{I}_n)\tilde{\mathbf{U}}_i\|_F\|(\mathbf{U}_i^\top\mathbf{U}_i+\lambda\mathbf{I}_r)^{-1}\|_2\\
		\leq&\|(\mathbf{X}+\lambda\mathbf{I}_n)\|_2\|\tilde{\mathbf{U}}_i\|_F\|(\tilde{\mathbf{U}}_i^\top\tilde{\mathbf{U}}_i+\lambda\mathbf{I}_r)^{-1}-(\mathbf{U}_i^     T\mathbf{U}_i+\lambda\mathbf{I}_r)^{-1}\|_2\\
		& + \|(\mathbf{X}+\lambda\mathbf{I}_n)\|_2\|\tilde{\mathbf{U}}_i\|_F\|(\mathbf{U}_i^\top\mathbf{U}_i+\lambda\mathbf{I}_r)^{-1}\|_2\epsilon\\
		\leq&[\frac{2(B+\lambda)a^2}{(\lambda-a^2)(\lambda-a^2-2a\epsilon)}+\frac{(B+\lambda)a}{\lambda-a^2}]\epsilon\\
		=&[\frac{2(B+\lambda)a^2}{(\lambda-a^2)(\lambda-a^2-\frac{\lambda}{2}+\frac{\lambda}{2}-2a\epsilon)}+\frac{(B+\lambda)a}{\lambda-a^2}]\epsilon.
		\end{aligned}
			\end{equation*}
		Then
		\begin{equation*}
		\begin{aligned}
			\|\tilde{\mathcal{F}}_{Ui}(\mathbf{U}_i)-\tilde{\mathcal{T}}_{Ui}(\tilde{\mathbf{U}}_i)\|_F \leq [\frac{4(B+\lambda)a^2}{(\lambda-a^2)^2}+\frac{(B+\lambda)a}{\lambda-a^2}]\epsilon
		\end{aligned}
		\end{equation*}
		Letting
		\begin{equation*}
		C =\frac{4(B+\lambda)a^2}{(\lambda-a^2)^2}+\frac{(B+\lambda)a}{\lambda-a^2}.
		\end{equation*}
		, with the continuity of the ReLU function, we proved that $\forall \mathbf{U}_i\in\mathcal{B}(\tilde{\mathbf{U}}_i,\epsilon)$,
		\begin{equation*}
		\|\mathcal{F}_{Ui}(\mathbf{U}_i)-\mathcal{T}_{Ui}(\tilde{\mathbf{U}_i})\|_F\leq C\epsilon.
		\end{equation*}
		which means that $\mathcal{F}_{Ui}$ is proximal. In this way, the $i$-th block is proximal to the original iteration. Since we choose $i$ randomly for the proof,  every block is $\epsilon$-proximal when the above conditions are satisfied. 
	\end{proof}